\documentclass[lettersize,journal]{IEEEtran}
\usepackage{amsmath,amsfonts}
\usepackage{algorithmic}
\usepackage{array}
\usepackage[caption=false,font=normalsize,labelfont=sf,textfont=sf]{subfig}
\usepackage{textcomp}
\usepackage{stfloats}
\usepackage{url}
\usepackage{verbatim}
\usepackage{graphicx}
\def\BibTeX{{\rm B\kern-.05em{\sc i\kern-.025em b}\kern-.08em
    T\kern-.1667em\lower.7ex\hbox{E}\kern-.125emX}}
\usepackage{balance}

\usepackage{pifont}
\usepackage{diagbox}
\usepackage{booktabs}
\usepackage{color}
\usepackage{float}
\usepackage{enumerate}

\begin{document}
\title{BSDM: Background Suppression Diffusion Model for Hyperspectral Anomaly Detection}
\author{Jitao Ma, Weiying Xie, \textit{Member, IEEE}, Yunsong Li, \textit{Member, IEEE}, Leyuan Fang, \textit{Senior Member, IEEE}
\thanks{Jitao Ma, Weiying Xie and Yunsong Li are with the State Key Laboratory of Integrated Services Networks, Xidian University, Xi'an 710071, China (e-mail: 21011210271@stu.xidian.edu.cn; wyxie@xidian.edu.cn; ysli@mail.xidian.edu.cn).

Leyuan Fang is with the College of Electrical and Information Engineering, Hunan University, Changsha 410082, China, and also with the Peng Cheng Laboratory, Shenzhen 518000, China (e-mail: fangleyuan@gmail.com).}}

\markboth{Submitted to IEEE Transactions on Image Processing}%
{How to Use the IEEEtran \LaTeX \ Templates}

\maketitle

\begin{abstract}
Hyperspectral anomaly detection (HAD) is widely used in Earth observation and deep space exploration. A major challenge for HAD is the complex background of the input hyperspectral images (HSIs), resulting in anomalies confused in the background. On the other hand, the lack of labeled samples for HSIs leads to poor generalization of existing HAD methods. This paper starts the first attempt to study a new and generalizable background learning problem without labeled samples. We present a novel solution BSDM (background suppression diffusion model) for HAD, which can simultaneously learn latent background distributions and generalize to different datasets for suppressing complex background. It is featured in three aspects: (1) For the complex background of HSIs, we design pseudo background noise and learn the potential background distribution in it with a diffusion model (DM). (2) For the generalizability problem, we apply a statistical offset module so that the BSDM adapts to datasets of different domains without labeling samples. (3) For achieving background suppression, we innovatively improve the inference process of DM by feeding the original HSIs into the denoising network, which removes the background as noise. Our work paves a new background suppression way for HAD that can improve HAD performance without the prerequisite of manually labeled data. Assessments and generalization experiments of four HAD methods on several real HSI datasets demonstrate the above three unique properties of the proposed method. The code is available at \url{https://github.com/majitao-xd/BSDM-HAD}.
\end{abstract}


\section{Introduction}
\IEEEPARstart{R}{ecently}, anomaly detection techniques are widely used in various fields, such as natural image anomaly detection \cite{4,5,6,tip:1,tip:2}, video anomaly detection \cite{1,2,3,tip:3,tip:4} and hyperspectral anomaly detection (HAD) \cite{7,8,9,tip:5}. Among them, HAD is one of the most important technologies for Earth observation and deep space exploration. It aims to detect anomalies from high-dimensional HSIs. Different from natural images with only three channels, HSI is a kind of images with tens to hundreds of channels, where each pixel is a spectral vector. In addition, HSI has complex and diverse ground objects with low resolution, which makes it difficult to handle with common anomaly detection methods. Therefore, it is highly essential to design anomaly detection methods specifically for HSI. 

\begin{figure}[t]
	\centering
	\includegraphics[width=0.5\textwidth]{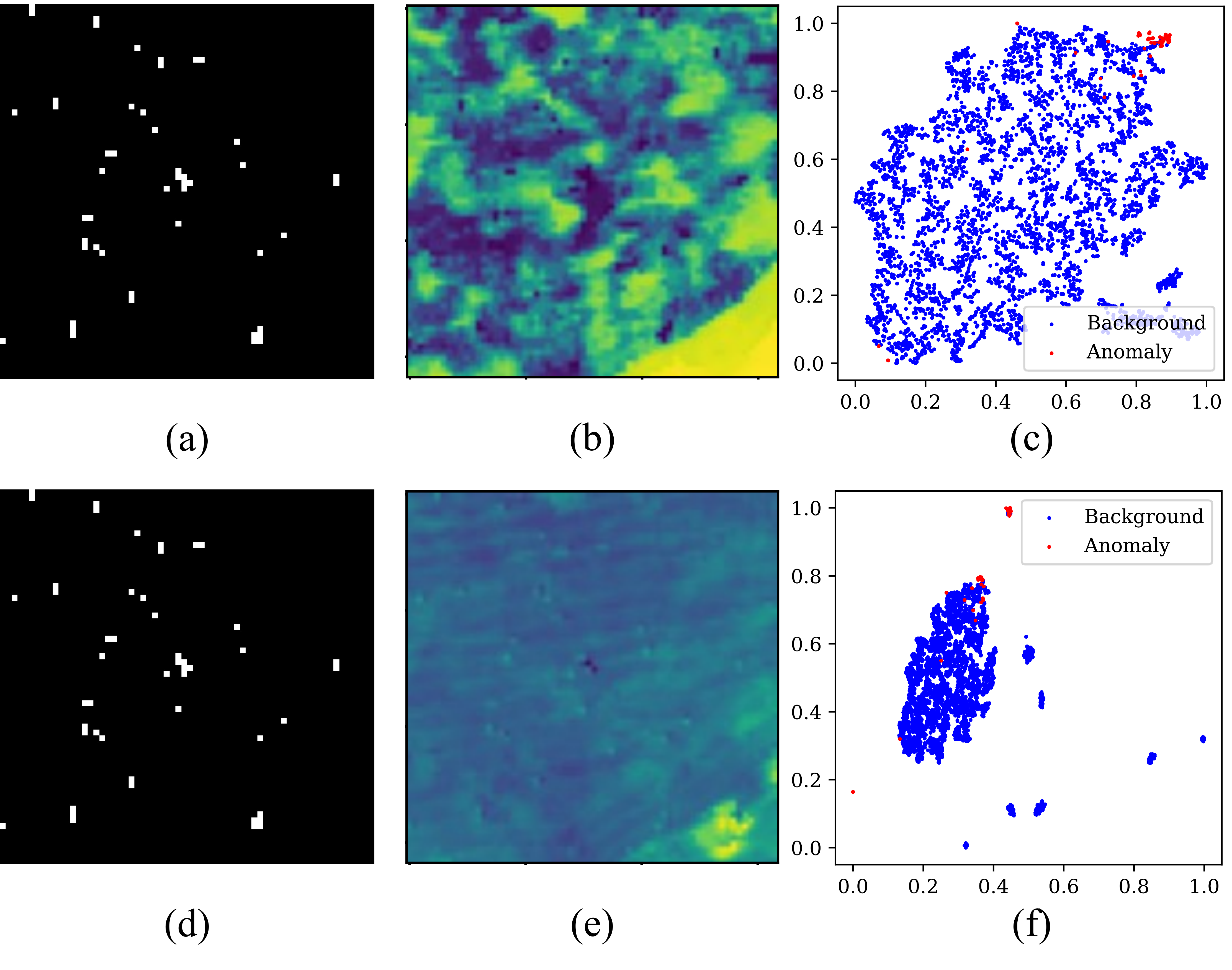}
	\caption{Visualization of the HAD-100 dataset and visualization of the HAD-100 dataset after background suppression by BSDM. (a) Anomalies map. (b) 1D t-SNE map. (c) 2D t-SNE scatter plot. (d) Anomalies map. (e) 1D t-SNE map after background suppression by BSDM. (f) 2D t-SNE scatter plot after BSDM suppression background.}
	\label{fig:t-sne}
\end{figure}

A series of model-driven approaches for HAD have been proposed. Many of them, such as RX \cite{rx}, LRX \cite{lrx}, CRD \cite{crd}, AED \cite{aed}, mathematically explore the differences between spectral vectors to determine the background and anomalies. Subsequently, with the development of deep learning, many HAD techniques based on deep neural networks (DNNs) emerged. The majority of these data-driven methods are based on networks such as AE \cite{ae}, AAE \cite{aae}, GAN \cite{gan} to detect anomalies mining the latent features or reconstruction errors of HSI. Theoretically, these methods can accurately locate anomalies. In practice, the majority of HSIs possess complex backgrounds, resulting in certain anomalies being confused within the background, as shown in Fig. \ref{fig:t-sne}. Likewise, this is the fundamental reason for the poor generalization of existing HAD methods. Consequently, it is difficult to obtain the desired detection results in real scenarios. There are also methods, such as LREN \cite{lren}, weaklyAD \cite{wad}, that consider this issue. However, these methods only improve performance by additional constraints and do not present a fundamental solution for HAD.

A natural question is: is it possible to develop a generalizable method to suppress the background of HSIs without labeling samples? In order to address this question, we systematically discuss various DNN architectures and propose our approach based on the following properties: 1) it does not require any manual labeling; 2) it has excellent generalization; 3) it takes only little time to compute. In this way, we propose a background suppression diffusion model (BSDM) for HAD. To the best of our knowledge, this is the first modular method for suppressing the background of HSI. Moreover, this is the first method to introduce diffusion model (DM) into HAD. DM \cite{ddpm} is a deep learning architecture that is widely used for tasks such as image generation \cite{10,11,12}, image reconstruction \cite{13,14,15}. In contrast to most of the existing DM-based methods, we focus more on the denoising and characterization performance of DM than the image generation performance.

The core idea of BSDM is to treat the background as noise that needs to be removed. Therefore, we take a pseudo background noise instead of the commonly used standard Gaussian noise. The pseudo background noise is obtained from input HSIs and thus does not require manual labeling. It motivates the network to learn the latent background distribution of input HSIs, so that BSDM can suppress the background as noise in the inference process. Meanwhile, we observe that despite the powerful characterization capability of DM, the complex background of HSIs makes it generalizable only on datasets with similar scenarios. To further improve the generalization of BSDM, we propose a statistical offset module that embeds the statistical values of input HSI into network. The statistical offset module reduces the feature distance between the source and target domains on the representation space by offsetting the distribution of feature maps for different input HSIs to the same region, so that BSDM can adapt to datasets from different domains. Most importantly, we innovatively improve the inference process of DM by feeding the original HSIs into the denoising network, so that the background is removed as noise. Notably, a well-trained BSDM has strong denoising ability, so in practice, the BSDM only needs to perform the inference process to suppress the background, and no longer needs to spend a lot of time on training. Extensive experiments demonstrate the effectiveness of BSDM in suppressing background and improving the performance of HAD methods. Our main contributions are summarized as follows.

\begin{itemize}
	\item As a pioneer work, we introduce diffusion models to HAD and propose a generalizable approach named BSDM for suppressing the background of HSI. This design can be applied as a pre-processing module to other HAD methods for improving their performances.
	\item To further improve the generalization of BSDM, we propose a statistical offset module that embeds the statistical values of input image into network, so that BSDM can adapt to datasets from different domains.
	\item Sufficient assessment and generalizability experiments on several real hyperspectral datasets and four HAD methods demonstrate that BSDM can effectively suppress the background and improve HAD performance in a variety of complex scenes.
\end{itemize}

\section{Related Works}
\subsection{Anomaly Detection}
Anomaly detection is an important technique in artificial intelligence for finding unusual observations whose characteristics are different from the primary data. Anomaly detection techniques have been applied to many fields \cite{16,17,18}. They are mostly trained on plenty of normal data or manually labeled anomalous data. Moreover, most of these methods are designed for natural images with simple backgrounds. Therefore, they have limited application on HSIs with complex backgrounds and lack of manual labeling. 

\subsection{Hyperspectral Anomaly Detection}
With respect to HAD, the existing approaches can be grouped into model-driven and data-driven. 
\subsubsection{Model-driven HAD}
Model-driven HAD methods mathematically explore the differences between spectral vectors to locate anomalies. For example, The RX algorithm builds a background model based on the assumption that the background obeys a Gaussian distribution, and calculates the Mahalanobis distance between the spectrum and the background model to determine the anomalous target. On this basis, the LRX algorithm applies the kernel density estimation method to estimate the background probability density function, and achieves local hyperspectral anomaly detection based on the background log-likelihood value. Since anomalous pixels in the above method pollute the background statistics, representation-based background models are proposed. CRD assumes that a background pixel can be linearly approximated by the pixels around it, and builds a background model by this decision. AED combines attribute filters and differential operations to remove the background. FrFE \cite{frfe} applies fractional-order Fourier transform as pre-processing to obtain features and introduces the Shannon entropy uncertainty principle to significantly distinguish the signal from background and noise. SSDF \cite{ssdf} combines dimensionality reduction and data splitting techniques with an isolation-based discriminative forest model designed to avoid background contamination caused by anomalous targets. 2S-GLRT \cite{glrt} detects anomalies with unknown spatial patterns in Gaussian background HSIs based on generalized likelihood ratio test. In conclusion, background modeling is crucial for model-driven HAD methods. However, as scenes and applications become more complex, the background of HSI becomes complicated, which limits the detection performance of these methods.

\subsubsection{Data-driven HAD}
Data-driven HAD methods are based on deep learning to detect anomalies by reconstruction or estimation. E2E-LIADE \cite{e2e} uses AE as a reconstruction network and fuses the reconstruction error space and extracted features into a density estimation network fitting a mixed Gaussian distribution. With end-to-end optimization, E2E-LIADE obtains the energy map as the detection result. Similarly, there are some other approaches using different generative networks, SAFL \cite{safl} employing adversarial AE to extract the features of the latent layer. HADGAN \cite{hadgan} proposes spectrally constrained GAN networks to reduce data dimensionality for background modeling. In addition, there are also methods to mine the features of HSIs from different dimensions. WeaklyAD takes GAN as the backbone and introduces a Kullback-Leible divergence-based orthogonal projection divergence spectral constraint to homogenize the background. DFAE \cite{dfae} decomposes HSIs into high-frequency and low-frequency components which are fed into different AEs. In recent years, some anomaly detection methods with transformer \cite{transformer} are proposed. Li \textit{et.al.} \cite{yoto} proposes a swin transformer-based \cite{swin-transformer} method that can learn the spatial context characteristics between anomalies and backgrounds in an unsupervised way. These methods can effectively catch most of the anomalies, however, the performance is not optimal because there is no maximum discrimination between background and anomalies. This issue is particularly serious for HSIs with anomalies confused in the background. The proposed BSDM is designed to solve this issue precisely by suppressing the background.

\subsection{Diffusion Models}
Diffusion models are a class of probabilistic denoising networks. Among them, denoising diffusion probabilistic models (DDPMs) \cite{ddpm} is the dominant diffusion model. DDPMs are divided into forward process (also known as diffusion process) and reverse process, which are both parameterized Markov chains. The forward process contains no learnable parameters, and this process adds a randomly generated standard Gaussian noise to the image at each time step. As the time step increases, the noise keeps diffusing in the original image, making the original image close to the standard Gaussian distribution. The purpose of the reverse process is to denoise and gradually recover the original image from the Gaussian noise, which is done by a neural network. The network used for denoising is usually U-Net \cite{19}, GAN \cite{20} or transformer \cite{21}. In practice, the stepwise noise addition of the forward process is reduced to direct noise addition according to a given time step. The reverse process first recovers the noise added in the forward process and then recovers the original image in combination with the noise-added image. Therefore, the noise added in the forward process is the known label, and the loss function used to train the denoising network in the inverse process is the $\ell_2$-norm between the noise before and after recovery. DDPMs are widely used due to their powerful generation, denoising and feature extraction capabilities. In recent years, there are also some methods introduced them into the field of remote sensing, such as HSIs classification \cite{22}, remote sensing change detection \cite{23}. However, to the best of our knowledge, very little work has been done to successfully apply DDPMs to HAD. DDPMs have strong denoising and characterization ability, and when applied to HAD, they can not only suppress the latent background distribution, but also increase the distance between anomalies and background, making anomalies easier to separate from the background.

\section{Method}
\subsection{Problem Formulation}
It is accepted that anomalies in HSIs have two distinctive characteristics compared to background: 1) the percentage of anomalies is small, usually no more than 2\% of all spectral vectors in HSI; 2) background can be represented by most other samples in the HSI, while anomalies cannot. Therefore, we can easily conclude that the distribution of input HSI is extremely close to the background distribution. BSDM takes this as a motivation to learn the latent background distribution and suppress the background. 

\begin{figure*}[ht]
	\centering
	\includegraphics[width=1.0\textwidth]{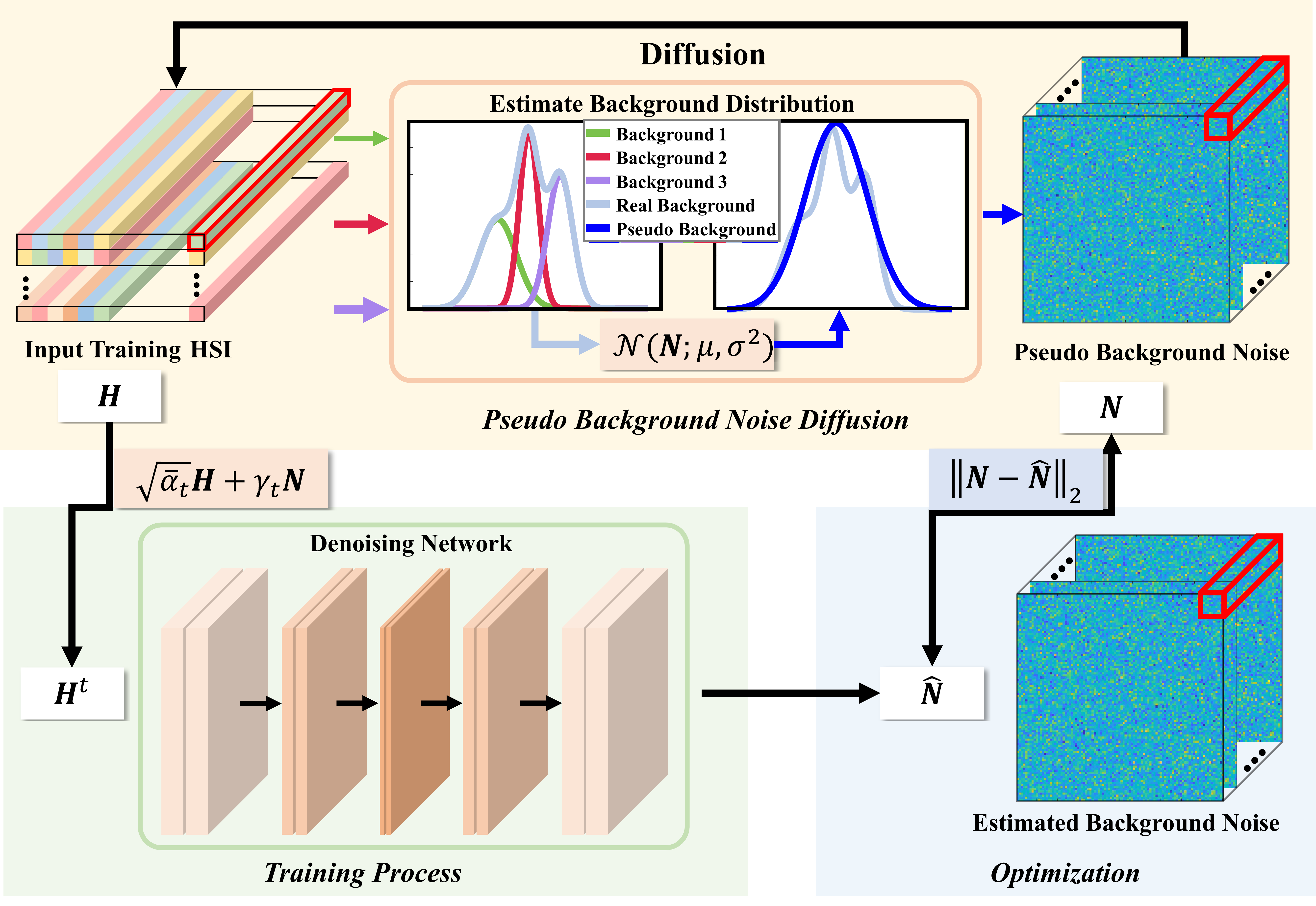}
	\caption{Overview diagram of the training process of BSDM. It estimates the distribution of the background before training to generate pseudo background noise $\boldsymbol{N}$. Although the real background distribution obeys a mixed Gaussian, the distribution of generated pseudo background noise is similar to the real background. BSDM is trained with the noise-diffused HSI $\boldsymbol{H}^t$ as input and outputs the estimated background noise $\boldsymbol{\hat{N}}$. Finally, the $\ell_2$-norm of the difference between $\boldsymbol{N}$ and $\boldsymbol{\hat{N}}$ is used as the loss function and optimized.}
	\label{fig:train}
\end{figure*}

We denote the input HSI by $\boldsymbol{H}=\left[\boldsymbol{h}_1, \ldots, \boldsymbol{h}_i, \ldots, \boldsymbol{h}_L\right] \in \mathbb{R}^{L \times B}$, where $\boldsymbol{h}_i,(i=1, \ldots, L)$ is the $i$-th spectral vector with $B$ bands in the original HSI space. First, we generate pseudo background noise $\boldsymbol{N}=\left[\boldsymbol{n}_1, \ldots, \boldsymbol{n}_i, \ldots, \boldsymbol{n}_L\right] \in \mathbb{R}^{L \times B}$ based on the input HSI, which has a distribution close to the background. As with the common DDPMs, $\boldsymbol{N}$ obeys Gaussian distribution. Therefore, we can still use the $\ell_2$-norm of input and output noise as the optimization target:
\begin{equation}
	\label{problem formulation}
	\begin{gathered}\min _{\theta, t}\left(\left\|\boldsymbol{N}-f\left(\boldsymbol{H}^t ; \theta\right)\right\|_2\right),
	\\ \text { s.t., } \boldsymbol{H}^t=\boldsymbol{H} \oplus_t \boldsymbol{N},\end{gathered}
\end{equation}
where $f\left(\boldsymbol{H}^t ; \theta\right)$ denotes the denoising network of BSDM, $\theta$ denotes the parameters of denoising network, $t$ denotes time step with interval $[1,T]$, $\left\| \cdot \right\|_2$ denotes the $\ell_2$-norm, $\boldsymbol{H}^t$ denotes the input image after $t$ times of noise diffusion and $\oplus_t$ denotes $t$ times of noise diffusion. Note that the use of pseudo background noise makes the noise diffusion equations of BSDM different from that of DDPMs, which are discussed in Section \ref{sec:pbn}. Since the high dimensionality of HSI, the use of randomly generated noise for each iteration leads to difficulties in convergence of the denoising network. BSDM uses the same noise and $t$ in all iterations to speed up convergence and shorten the training time, as shown in Fig. \ref{fig:train}. Finally, a well-trained denoising network treats the background distribution as noise.

It is worth noting that BSDM innovatively improves the inference process by feeding the original HSIs directly into the denoising network, as shown in Fig. \ref{fig:test}, that is, the denoising network removes the background as noise, as shown in the Fig. \ref{fig:t-sne}. We next discuss in detail suppressing the background of HSIs via BSDM.

\subsection{Pseudo Background Noise}
\label{sec:pbn}
Instead of the standard Gaussian noise used in the common DM, we diffuse the pseudo-background noise $\boldsymbol{N}$, which is closer to the background distribution, into the original HSI. In this way, it is much simpler for denoising network to extract noise from the background than from anomalies while training. As a result, the background is extracted as noise in the inference process. This is necessary to enhance the discrimination between backgrounds and anomalies.

In order to make the pseudo background noise closer to the original HSI, we consider various noise generation strategies, such as Gaussian distribution, mixed Gaussian distribution and Poisson distribution. Among them, the mixed Gaussian distribution has the strongest fit, followed by the Gaussian distribution, and the Poisson distribution is the worst. However, the mixed Gaussian distribution fits anomalies equally well, which can lead to overfitting of the denoising network and damage the generalizability of BSDM. Therefore, we use a Gaussian distribution with the same mean and variance to the input HSI as the pseudo background noise:
\begin{equation}
	\label{noise distribution}
	\boldsymbol{N} \sim \mathcal{N}(\mu, \sigma^2),
\end{equation}
where $\mu$ and $\sigma$ denote the mean and standard deviation of $\boldsymbol{H}$, respectively. Since the pseudo background noise does not obey the standard Gaussian distribution, the noise diffusion equations of DDPMs are not applicable to BSDM. Therefore, we design a specific noise diffusion method to deal with the proposed pseudo background noise as follows.

Given a data point $x_0$ sampled from the real data distribution $q(x)$, $(x_0 \sim q(x))$, the forward diffusion process can be defined as adding noise $\epsilon_0, \ldots, \epsilon_{t-2}, \epsilon_{t-1} \sim \mathcal{N}(\mu, \sigma^2)$ stepwise to $x_0$. This diffusion process can be formulated as follows:
\begin{equation}
	\label{diff:1}
	\begin{gathered}
	q(x_t \mid x_{t-1}) = \mathcal{N}(x_t; \sqrt{1 - \beta_t}x_{t-1} + \mu \sqrt{\beta_t}, \sigma^2 \beta_t \mathbf{I}),
	\\ 
	\text { s.t., } \beta_t = \frac{\lambda t}{T},
	\end{gathered}
\end{equation}
where $x_t$ denotes the data after $t$-step diffusion, $\beta_t$ denotes the noise intensity, $\lambda$ is a hyperparameter to limit the magnitude of $\beta_t$, and $\mathbf{I}$ is the identity matrix. Note that $q(x_t \mid x_{t-1})$ still obeys the Gaussian distribution. Thus, the diffusion process from step 1 to $T$ can be mathematically defined as:
\begin{equation}
	\label{diff:2}
	q(x_{1:T} \mid x_0) = \prod_{t=1}^{T}q(x_t \mid x_{t-1}),
\end{equation}
this is also the posterior probability. Similar to DDPMs, in order to make the above formulas computable, we apply the reparameterization trick to design a treatable closed-form sampling at any timestep. First, we define $\alpha_t = 1 - \beta_t$, $\overline{\alpha}_t = \prod_{i=1}^{t}\alpha_i$. In this way, $x_t$ can be calculated as:
\begin{equation}
	\label{diff:3}
	x_t = \sqrt{\alpha_t}x_{t-1} + \sqrt{\beta_t}\epsilon_{t-1}.
\end{equation}

By substituting $x_{t-1}$ into Equation \ref{diff:3}, we have:
\begin{equation}
	\label{diff:4}
	x_t = \sqrt{\alpha_t \alpha_{t-1}}x_{t-2} + \sqrt{\alpha_{t-1} \beta_t}\epsilon_{t-2} + \sqrt{\beta_t}\epsilon_{t-1}.
\end{equation}

Since all time steps have the same Gaussian noise, we only use $\epsilon$ to denote the noise in the subsequent derivations:
\begin{equation}
	\label{diff:5}
	x_t = \sqrt{\alpha_t \alpha_{t-1}}x_{t-2} + (\sqrt{\alpha_{t-1} \beta_t} + \sqrt{\beta_t})\epsilon.
\end{equation}

By deriving Equation \ref{diff:5} to $t=1$, we have:
\begin{equation}
	\label{diff:6}
	\begin{gathered}
		x_t = \sqrt{\overline{\alpha}_t}x_0 + \gamma_t \epsilon,
		\\ 
		\text { s.t., } \gamma_t \epsilon \sim \mathcal{N}(\mu \gamma_t, (1 - \overline{\alpha}_t)\sigma^2),
	\end{gathered}
\end{equation}
where $\gamma_t$ is defined as:
\begin{equation}
	\label{diff:7}
	\gamma_t = \sum_{k=0}^{t-1}\sqrt{\frac{\overline{\alpha}_t\beta_{t-k}}{\overline{\alpha}_{t-k}}}.
\end{equation}

Thus, Equation \ref{diff:2} can be simplified to the distribution of $x_t$:
\begin{equation}
	\label{diff:8}
	x_t \sim q(x_t \mid x_0) = \mathcal{N}(x_t; \sqrt{\overline{\alpha}_t}x_0 + \mu \gamma_t, (1 - \sqrt{\overline{\alpha}_t})\sigma^2 \mathbf{I}).
\end{equation}

Finally, the symbol $\oplus_t$ in Equation \ref{problem formulation} can be denoted mathematically as:
\begin{equation}
	\label{diff:9}
	\boldsymbol{H}^t = \boldsymbol{H} \oplus_t \boldsymbol{N} = \sqrt{\overline{\alpha}_t}\boldsymbol{H} + \gamma_t \boldsymbol{N}.
\end{equation}

\begin{figure}[ht]
	\centering
	\includegraphics[width=0.5\textwidth]{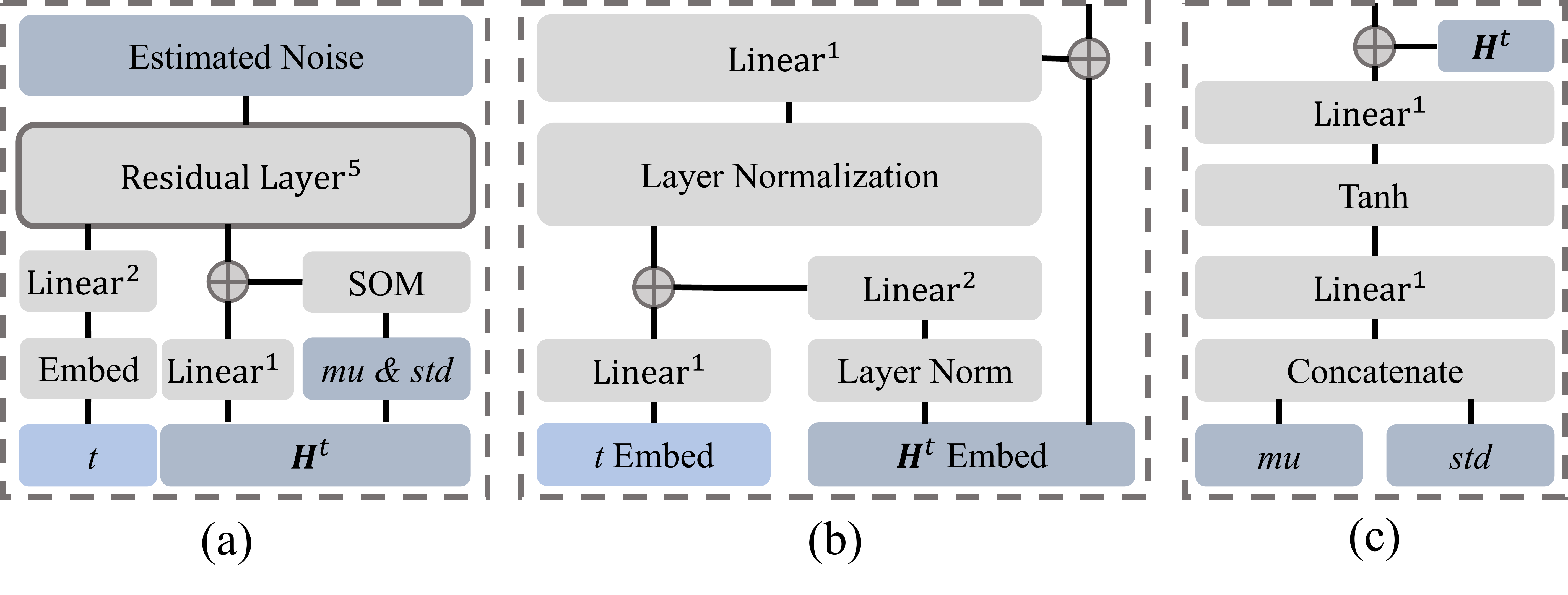}
	\caption{Network structure of (a) denoising network, (b) residual layer, (c) statistical offset module, where SOM denotes statistical offset module and $\mathrm{Linear}^i$ denotes $i$ cascading fully connected layers.}
	\label{fig:network}
\end{figure}

\subsection{Denoising Network}
We show the structure of the denoising network in Fig. \ref{fig:network} (a). It takes the spectral vector $\boldsymbol{h}_i^t$ in $\boldsymbol{H}^t$ as an input unit and the $i$-th vector $\hat{\boldsymbol{n}}_i$ of estimated noise $\hat{\boldsymbol{N}}$ as output:
\begin{equation}
	\label{train-input}
	\hat{\boldsymbol{n}}_i = f(\boldsymbol{h}_i^t; \theta).
\end{equation}

In deep learning based HAD methods, most of the networks are designed to be simple (typically only 5 fully connected layers) because of the limited available training sets. However, in DDPMs, a large number of trainable parameters are required for fitting the inverse process to estimate the noise. Therefore, we design a fully connected block with skip connect, called residual layer, as shown in Fig. \ref{fig:network} (b). This design can increase trainable parameters while avoiding gradient disappearance due to over-depth network. In addition, the embedding method for $t$ is the same as for common DDPMs:
\begin{equation}
	\label{t-embed}
	\begin{gathered}
		\hat{t}=\operatorname{Emb}_t\left(t ; \theta_{t e}\right), \\
		\text { s.t., } \theta_{t e} \in \theta,
	\end{gathered}
\end{equation}
where $\operatorname{Emb}_t(\cdot;\theta_{te})$ denotes the time step embedding network, $\theta_{te}$ denotes the parameter of $\operatorname{Emb}_t$.

\begin{figure}[ht]
	\centering
	\includegraphics[width=0.5\textwidth]{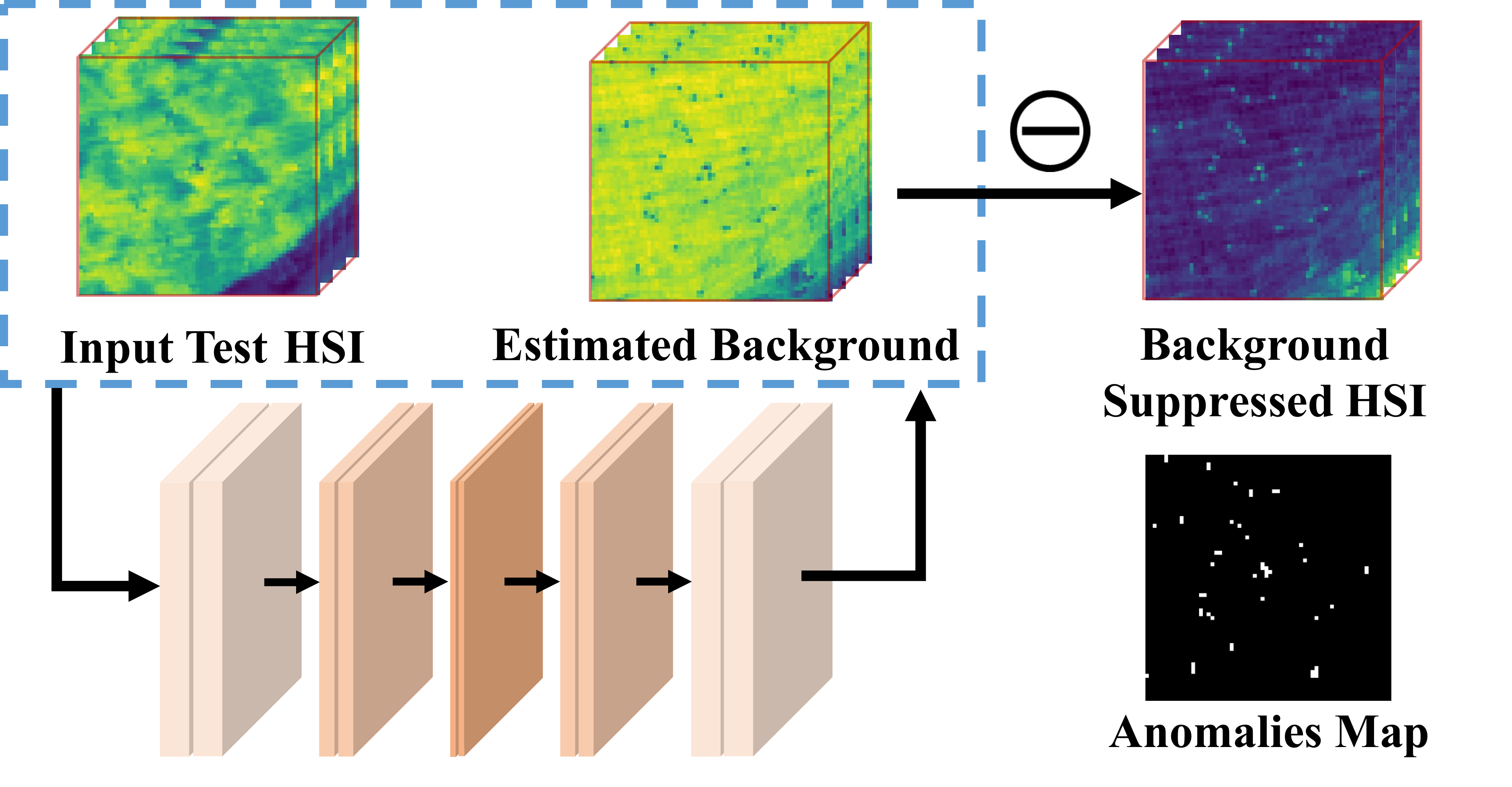}
	\caption{Overview diagram of the inference process of BSDM. It takes the test HSI without any preprocessing as input and outputs the estimated background. Finally, the original test HSI removes the estimated background to get the background suppressed HSI.}
	\label{fig:test}
\end{figure}

\subsection{Statistical Offset Module}
\label{sec:som}
\begin{figure}[t]
	\centering
	\includegraphics[width=0.5\textwidth]{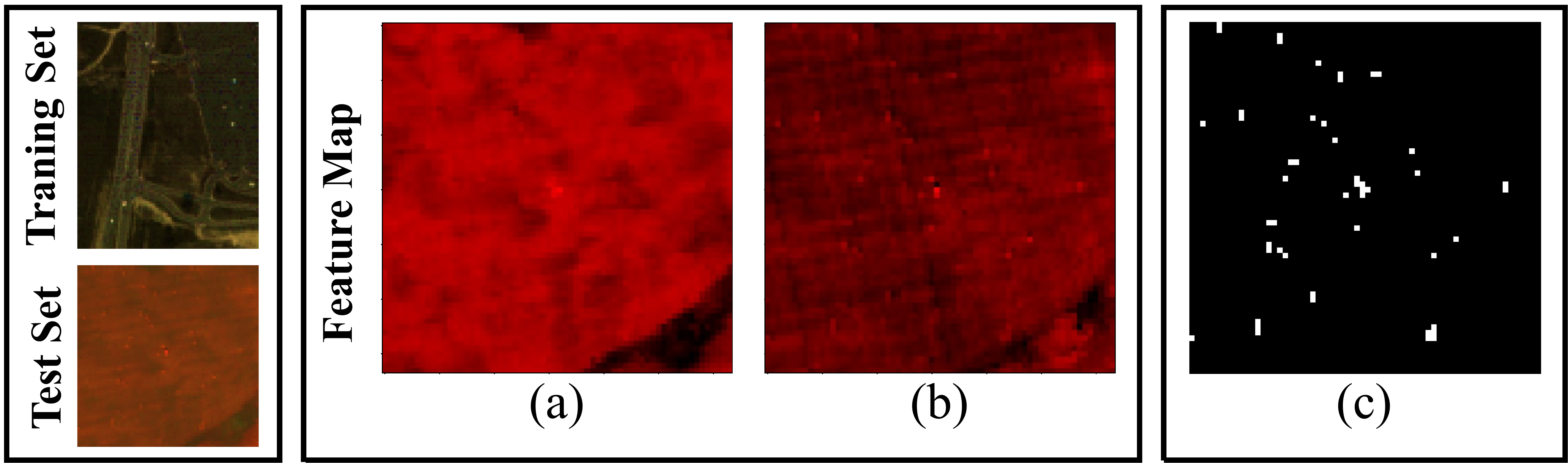}
	\caption{Visualization of the feature map during inference with Hydice as the training set and HAD-100 as the test set. (a) Without statistical offset module. (b) With statistical offset module. (c) Anomalies of HAD-100 for comparing the performance of feature maps to retain anomalous structures.}
	\label{fig:fea-map}
\end{figure}

Since $\mu$ and $\sigma$ are intrinsic to the input HSI, BSDM can learn the background distribution to some extent. However, the architecture of the denoising network motivates it to focus on learning feature extraction so that it can better reconstruct the input HSI. In contrast, the proposed statistical offset module motivates the denoising network to learn the distribution of the background so that it can better remove the noise. Although both may yield the same output in the training process, the latter allows the denoising network to adapt to different domain datasets in the inference process, as shown in Fig. \ref{fig:fea-map}.

As shown in Fig. \ref{fig:network} (c), the statistical offset module first calculates the mean and standard deviation of $\boldsymbol{h}_i^t$, and concatenates them:
\begin{equation}
	\label{get-s}
	\boldsymbol{s}_i^t = [\operatorname{E}[\boldsymbol{h}_i^t], \sqrt{\operatorname{D}[\boldsymbol{h}_i^t]}],
\end{equation}
where $\operatorname{E}[\cdot]$ and $\operatorname{D}[\cdot]$ denote calculation of mean and variance, respectively. Subsequently, $\boldsymbol{s}_i^t$ is input to the statistical offset module consisting of multiple fully connected layers:
\begin{equation}
	\label{s-embed}
	\begin{gathered}
		\hat{\boldsymbol{s}}_i^t = \operatorname{Emb}_s\left(\boldsymbol{s}_i^t ; \theta_{se}\right), \\
		\text { s.t., } \theta_{se} \in \theta,
	\end{gathered}
\end{equation}
where $\operatorname{Emb}_s(\cdot;\theta_{se})$ denotes the statistical offset module, $\theta_{se}$ denotes the parameter of $\operatorname{Emb}_s$. During the inference process, once the HSI from different domains is input, BSDM adapts its feature map distribution to the same region as training HSI.

\subsection{Inference Process}
The training process of BSDM is similar to that of common DDPMs, as shown in the Fig. \ref{fig:test}. The inference process is the core of background suppression. Since the noise extracted by the denoising network during the training process is similar to the distribution of the background. A well-trained denoising network is more likely to extract noise from spectral vectors with a distribution similar to that of the pseudo-background noise, that is, the denoising network is easier to suppress the intensity of the background. The inference process is represented mathematically as follows:
\begin{equation}
	\label{test}
	\hat{\boldsymbol{N}} = f(\boldsymbol{H};\theta),
\end{equation}
\begin{equation}
	\label{background-suppress}
	\boldsymbol{H}_{BS} = \boldsymbol{H} \ominus_t \hat{\boldsymbol{N}},
\end{equation}
where $\boldsymbol{H}_{BS}$ denotes the background suppressed HSI, $\ominus_t$ denotes minus based on time step $t$, which is the same as in the common DDPMs. Based on Equation \ref{diff:9}, $\ominus_t$ can be denoted mathematically as:
\begin{equation}
	\label{diff:10}
	\boldsymbol{H}_{BS} = \boldsymbol{H} \ominus_t \hat{\boldsymbol{N}} = \frac{1}{\sqrt{\overline{\alpha}_t}}(\boldsymbol{H} - \gamma_t \hat{\boldsymbol{N}}),
\end{equation}

\begin{figure}[!h]
	\centering
	\includegraphics[width=0.5\textwidth]{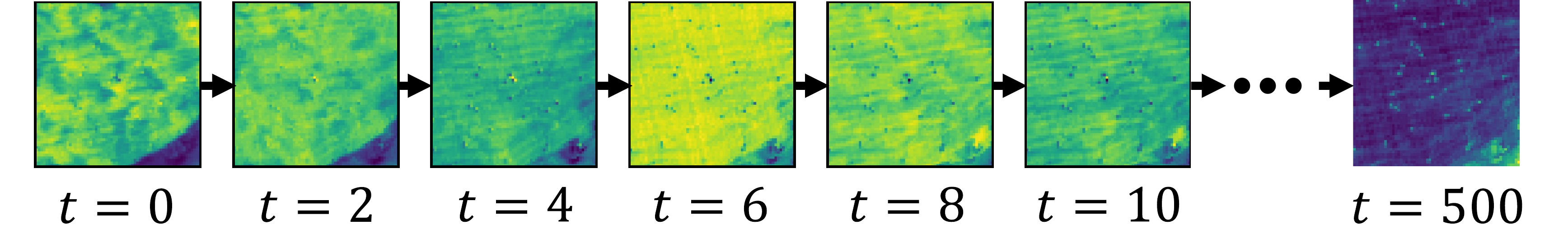}
	\caption{Visualization of the sample of the BSDM background suppression process. From $t$ = 0, the background is gradually suppressed until the result obtained by Equation \ref{diff:10}. Note that in practice the background is suppressed in one step, not gradually.}
	\label{fig:sample-test}
\end{figure}

In DDPMs, the inverse process is sampled in order to demonstrate denoising or generation performance. Thus, to demonstrate the performance of BSDM background suppression, we sample and visualize the background suppression process, as shown in Fig. \ref{fig:sample-test}.

More importantly, BSDM can suppress background more significantly by multiple inference. Theoretically, the intensity of the background can be suppressed to 0 after multiple inferences. However, since the distribution of the pseudo background noise is not exactly the same as the background, trace anomalies are also suppressed during the inference process. Therefore, BSDM cannot suppress the background without limit. Through extensive statistical experiments, we find that multiple BSDM inferences improve the performance of the model-driven HAD methods, while too many inferences can lead to unstable performance of data-driven HAD methods. We conjecture that this is caused by the randomness of DNNs.

In addition, HSIs acquired by different sensors have different band numbers, but the denoising network requires the test HSIs to have the same band number as the training HSIs. This issue prevents some of the test HSIs from being fed into the denoising network. To solve this problem, we design a band alignment strategy so that the number of bands of the test HSIs is the same as the training HSIs. When the number of bands $B_{test}$ of the test HSIs is smaller than the number of bands $B_{train}$ of the training HSIs, the test HSIs expand the bands by mirroring with the last band as the center. When $B_{test}>B_{train}$, the test HSIs randomly remove a band until $B_{test}=B_{train}$. Note that when $B_{test}>2\times B_{train}$, we first divide the test set from the band dimension into multiple batches for inference in turn, and concatenate the outputs, which avoids excessive loss of spectral features. 

\section{Experiments}
\subsection{Experimental Settings}
\subsubsection{Datasets}
We validate BSDM on six HSIs of different scenarios, including SanDiego \cite{sandiego}, Hydice \cite{crd}, Airport-Beach-Urban \cite{aed} and HAD-100 \cite{yoto}, which are listed as follows. 
\begin{enumerate}[(i)]
	\item \textit{SanDiego dataset:} This dataset was captured by the Airborne Visible/ Infrared Imaging Spectrometer (AVIRIS) sensor over the San Diego airport area, CA, USA. This image has 100$\times$100 pixels and 189 bands.
	\item \textit{Hydice dataset:} This dateset was captured by the Hyperspectral Digital Imagery Collection Experiment sensor over an urban area, CA, USA. This image has 80$\times$100 pixels and 162 bands.
	\item \textit{Airport-Beach-Urban:} This dateset was captured by the AVIRIS sensor. This dataset contains multiple HSIs, we choose three of them (Airport-1, Beach-1, Urban-3, which are denoted by Airport, Beach, Urban, respectively). Airport has 100$\times$100 pixels and 205 bands. Beach has 150$\times$150 pixels and 188 bands. Urban has 100$\times$100 pixels and 191 bands. 
	\item \textit{HAD-100 dataset:} This dateset was captured by the AVIRIS sensor. This dataset contains multiple HSIs, we choose one of them (HAD-20170821T183707\_96, which is denoted by HAD-100). HAD-100 has 64$\times$64 pixels and 425 bands. 
\end{enumerate}

Note that these datasets contain both point and structure anomalies of different sizes. Meanwhile, these datasets differ in sensors, capture time, and wavelength ranges.

\begin{figure*}[ht]
	\centering
	\includegraphics[width=1.0\textwidth]{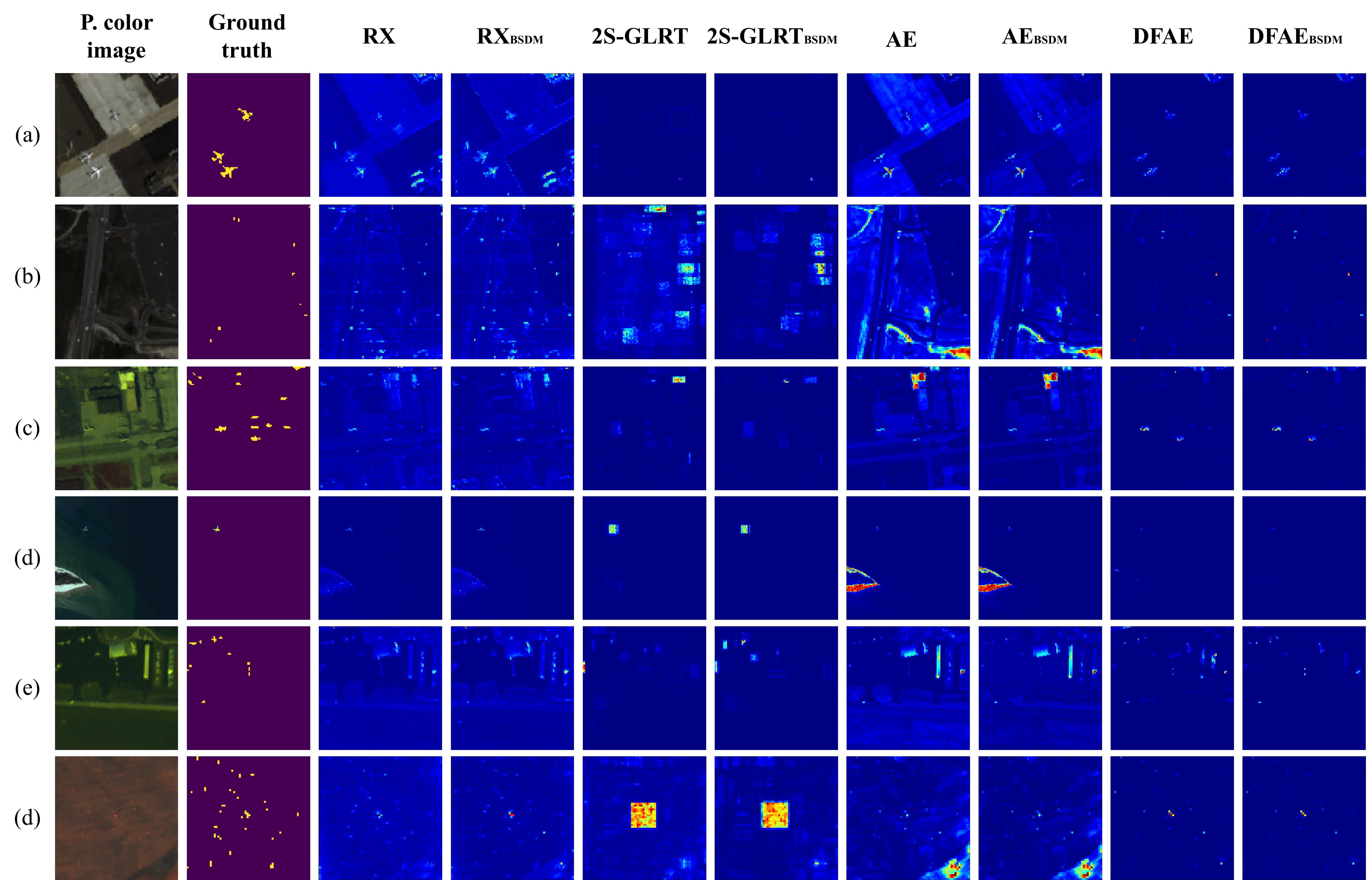}
	\caption{Pseudo color images, ground truth (Anomalies maps), and detection maps for (a) SanDiego, (b) Hydice, (c) Airport, (d) Beach, (e) Urban, (f) HAD-100. BSDM in the lower corner indicates the detection results after background suppression pre-processing by BSDM.}
	\label{fig:result-map}
\end{figure*}

\subsubsection{HAD Methods}
We validated the background suppression ability of BSDM on 4 HAD methods including RX, 2S-GLRT, AE, DFAE. RX is the classical model-driven HAD method that searches for anomalies by computing the Mahalanobis distance between samples. 2S-GLRT is a model-driven HAD method based on the generalized likelihood ratio test design method and ad hoc modification of it. Output and input windows are set to [19,15]. AE is a data-driven HAD method that takes the mean square error (MSE) before and after autoencoder reconstruction as the detection result. The hidden layer dimension of the autoencoder is set to [100,70,50,70,100]. DFAE is a data-driven HAD method that converts input HSIs into high-frequency components and low-frequency components. DFAE uses the parameter settings of the original paper. Note that these methods contain both classical and state-of-the-art, model-driven and data-driven.

\subsubsection{Evaluation Metrics}
To quantitatively validate the performance of BSDM, the receiver operating characteristic curve (ROC) together with the area under it are used as metrics. The area under the ROC curve of $(P_d,P_f)$ demonstrates the detection accuracy, and the closer this value is to 1, the better the detection capability. The area under the ROC curve of $(P_f,\tau)$ demonstrates the false alarm rate, and the closer this value is to 0, the better the false alarm rate. Where, $P_d$, $P_f$ and $\tau$ denote the true positive rate, false positive rate and threshold, respectively. Box-Whisker Plot \cite{box} is used to indicate the discrimination between background and anomalies.

\subsubsection{Experimental Environment}
The denoising network structure of BSDM is shown in Table \ref{tab:network}. We train BSDM with Adam optimizer, which has an initial learning rate of 1e-4 and decreases to 1e-5 with cosine scheduler. Batch size is set to the number of pixels input. The hyperparameter $\lambda$ is set to 0.02. The inference count of BSDM is set to $K$. $K$ and $t$ are set separately in different HAD method. $K$ and $t$ for RX, 2S-GLRT, AE, DFAE are set to [10, 500], [10, 40], [1, 30], [1, 30] respectively. We train the BSDM for 500 epochs by PyTorch on one NVIDIA 3090 Ti GPU with 24 GB memory.

\begin{table}[t]
	\centering
	\caption{Detailed Parameter Settings of the Denoising Network. Activ. Denotes the Activation Function}
	\begin{tabular}{c|c|c|c}
		\toprule[0.5mm]
		\textbf{Module} & \textbf{Base layer} & \textbf{Hidden dim} & \textbf{Activ.} \\ \midrule
		\textbf{Denoising network} & Residual layer & 100,50,50,100 & - \\ \midrule
		\textbf{Residual layer} & Linear layer & 200 & Tanh \\ \midrule
		\textbf{Time embedding layer} & Linear layer & 512 & Tanh \\ \midrule
		\textbf{Statistical offset module} & Linear layer & 512 & Tanh \\ \bottomrule[0.5mm]
	\end{tabular}
	\label{tab:network}
\end{table}

\begin{figure*}[ht]
	\centering
	\includegraphics[width=0.95\textwidth]{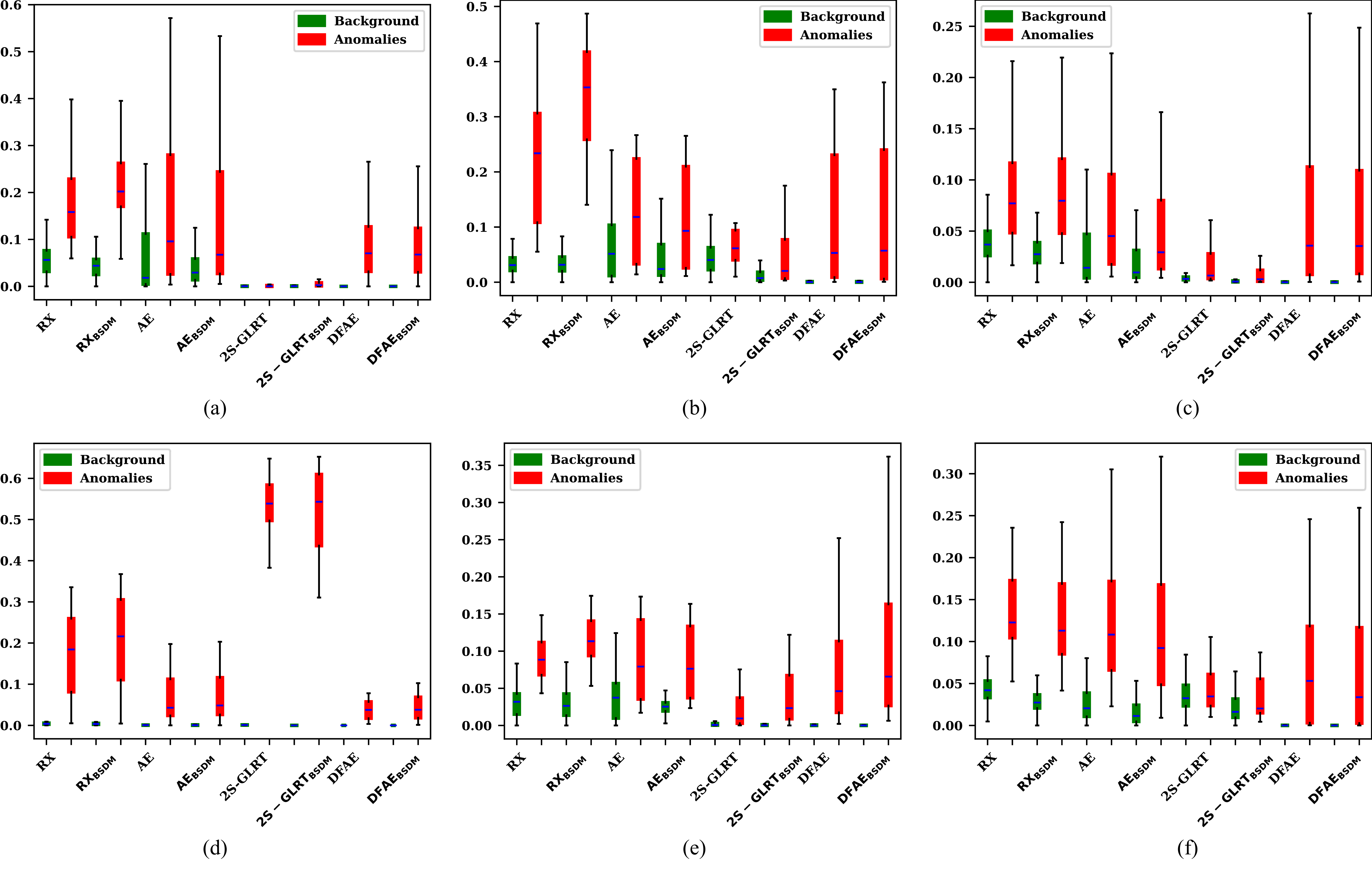}
	\caption{Anomaly and background separation maps visualized with Box-Whisker Plot on (a) SanDiego, (b) Hydice, (c) Airport, (d) Beach, (e) Urban, (f) HAD-100.}
	\label{fig:box}
\end{figure*}

\begin{table*}[ht]
	\centering
	\caption{Values of the Area Under the ROC $(P_d,P_f )$ for Different Methods on 6 Datasets. The Higher of This Value, the Higher the Accuracy.}
	\begin{tabular}{c||cc||cc||cc||cc}
		\toprule[0.5mm]
		\textbf{Method} & \multicolumn{2}{c||}{\textbf{RX}} & \multicolumn{2}{c||}{\textbf{2S-GLRT}} & \multicolumn{2}{c||}{\textbf{AE}} & \multicolumn{2}{c}{\textbf{DFAE}} \\ \midrule
		\textbf{BSDM} & \ding{55} & \ding{51} & \ding{55} & \ding{51} & \ding{55} & \ding{51} & \ding{55} & \ding{51} \\ \midrule
		\textbf{SanDiego} & 0.9403 & 0.9742 ($\uparrow$ 0.0339) & 0.5918  & 0.8557 ($\uparrow$ 0.2639) & 0.6730  & 0.6979 ($\uparrow$ 0.0249) & 0.9791  & 0.9799 ($\uparrow$ 0.0008) \\ 
		\textbf{Hydice} & 0.9763 & 0.9950 ($\uparrow$ 0.0187) & 0.6533  & 0.6928 ($\uparrow$ 0.0395) & 0.6844  & 0.7052 ($\uparrow$ 0.0208) & 0.9395  & 0.9426 ($\uparrow$ 0.0031) \\ 
		\textbf{Airport} & 0.8221 & 0.8814 ($\uparrow$ 0.0593) & 0.7288  & 0.7724 ($\uparrow$ 0.0436) & 0.7494  & 0.7519 ($\uparrow$ 0.0025) & 0.9693  & 0.9724 ($\uparrow$ 0.0031) \\ 
		\textbf{Beach} & 0.9808 & 0.9819 ($\uparrow$ 0.0009) & 0.9990  & 0.9990 ({--} 0.0000) & 0.9182  & 0.9186 ($\uparrow$ 0.0004) & 0.9973  & 0.9994 ($\uparrow$ 0.0021) \\ 
		\textbf{Urban} & 0.9513 & 0.9655 ($\uparrow$ 0.0142) & 0.8446  & 0.9119 ($\uparrow$ 0.0673) & 0.8249  & 0.8446 ($\uparrow$ 0.0197) & 0.9706  & 0.9942 ($\uparrow$ 0.0236) \\ 
		\textbf{HAD-100} & 0.9797 & 0.9801 ($\uparrow$ 0.0004) & 0.5344  & 0.6034 ($\uparrow$ 0.0690) & 0.8881  & 0.9095 ($\uparrow$ 0.0214) & 0.9890  & 0.9892 ($\uparrow$ 0.0002) \\ \bottomrule[0.5mm]
	\end{tabular}
	\label{tab:results-auc}
\end{table*}

\begin{table*}[ht]
	\centering
	\caption{Values of the Area Under the ROC $(P_f,\tau)$ for Different Methods on 6 Datasets. The Lower of This Value, the Better the False Alarm Rate.}
	\begin{tabular}{c||cc||cc||cc||cc}
		\toprule[0.5mm]
		\textbf{Method} & \multicolumn{2}{c||}{\textbf{RX}} & \multicolumn{2}{c||}{\textbf{2S-GLRT}} & \multicolumn{2}{c||}{\textbf{AE}} & \multicolumn{2}{c}{\textbf{DFAE}} \\ \midrule
		\textbf{BSDM} & \ding{55} & \ding{51} & \ding{55} & \ding{51} & \ding{55} & \ding{51} & \ding{55} & \ding{51} \\ \midrule
		\textbf{SanDiego} & 0.0589 & 0.0457 ($\downarrow$ 0.0132) & 0.0012  & 0.0011 ($\downarrow$ 0.0001) & 0.0587  & 0.0422 ($\downarrow$ 0.0165) & 0.0021  & 0.0019 ($\downarrow$ 0.0002) \\ 
		\textbf{Hydice} & 0.0380 & 0.0274 ($\downarrow$ 0.0106) & 0.0586  & 0.0240 ($\downarrow$ 0.0346) & 0.0767  & 0.0575 ($\downarrow$ 0.0192) & 0.0021  & 0.0019 ($\downarrow$ 0.0002) \\ 
		\textbf{Airport} & 0.0423 & 0.0336 ($\downarrow$ 0.0087) & 0.0088  & 0.0037 ($\downarrow$ 0.0051) & 0.0338  & 0.0291 ($\downarrow$ 0.0047) & 0.0018  & 0.0018 ({--} 0.0000) \\ 
		\textbf{Beach} & 0.0065 & 0.0064 ($\downarrow$ 0.0001) & 0.0036  & 0.0016 ($\downarrow$ 0.0020) & 0.0141  & 0.0140 ($\downarrow$ 0.0001) & 0.0003  & 0.0001 ($\downarrow$ 0.0002) \\ 
		\textbf{Urban} & 0.0350 & 0.0350 ({--} 0.0000) & 0.0051  & 0.0048 ($\downarrow$ 0.0003) & 0.0377  & 0.0348 ($\downarrow$ 0.0029) & 0.0046  & 0.0014 ($\downarrow$ 0.0032) \\ 
		\textbf{HAD-100} & 0.0455 & 0.0342 ($\downarrow$ 0.0113) & 0.0649  & 0.0526 ($\downarrow$ 0.0123) & 0.0390  & 0.0315 ($\downarrow$ 0.0075) & 0.0019  & 0.0018 ($\downarrow$ 0.0001) \\ \bottomrule[0.5mm]
	\end{tabular}
	\label{tab:results-fpr}
\end{table*}

\begin{figure*}[ht]
	\centering
	\includegraphics[width=1.0\textwidth]{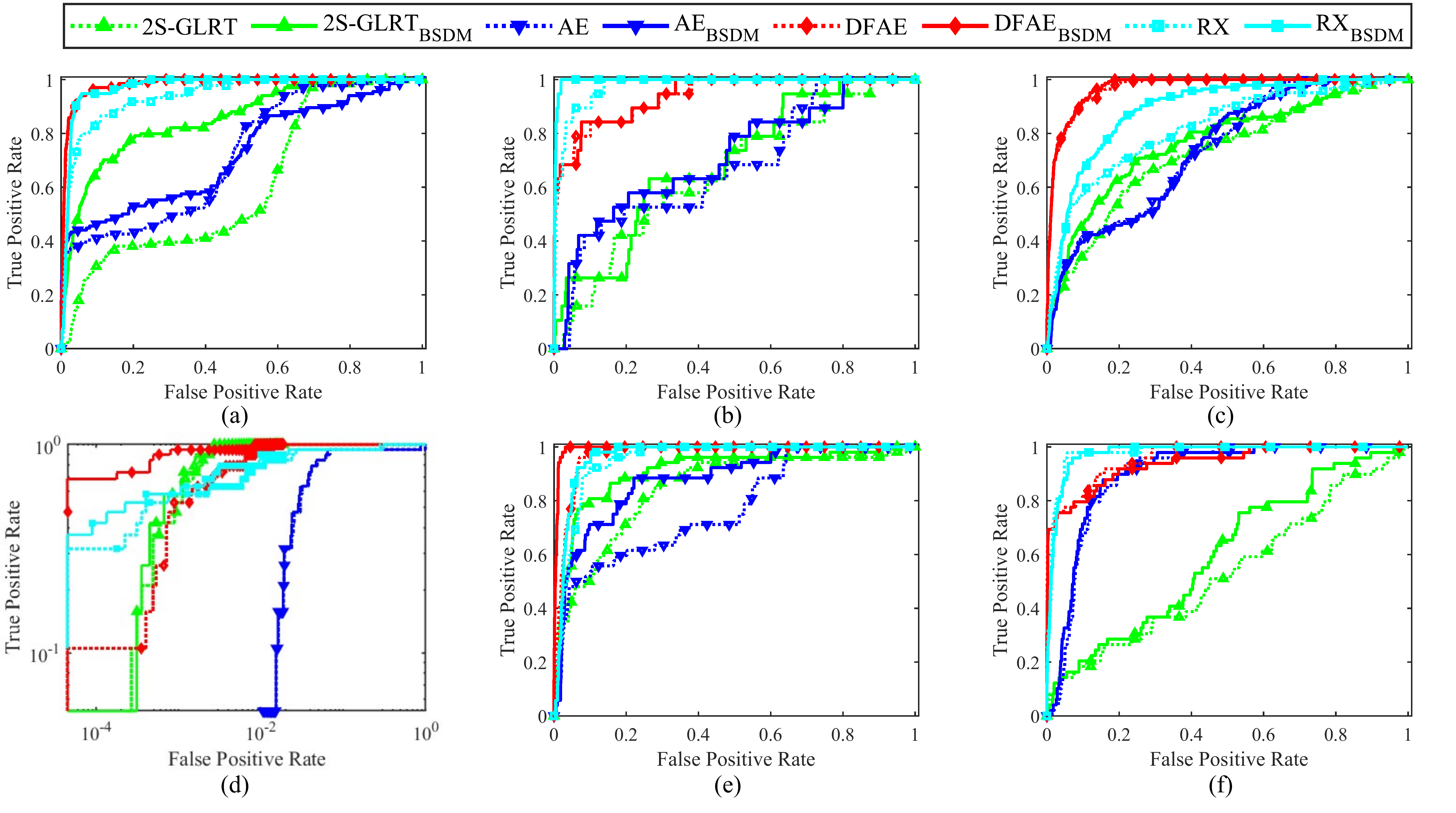}
	\caption{ROC curves of $(P_d,P_f)$ for different methods on (a) SanDiego, (b) Hydice, (c) Airport, (d) Beach, (e) Urban, (f) HAD-100.}
	\label{fig:roc}
\end{figure*}

\begin{figure*}[ht]
	\centering
	\includegraphics[width=1.0\textwidth]{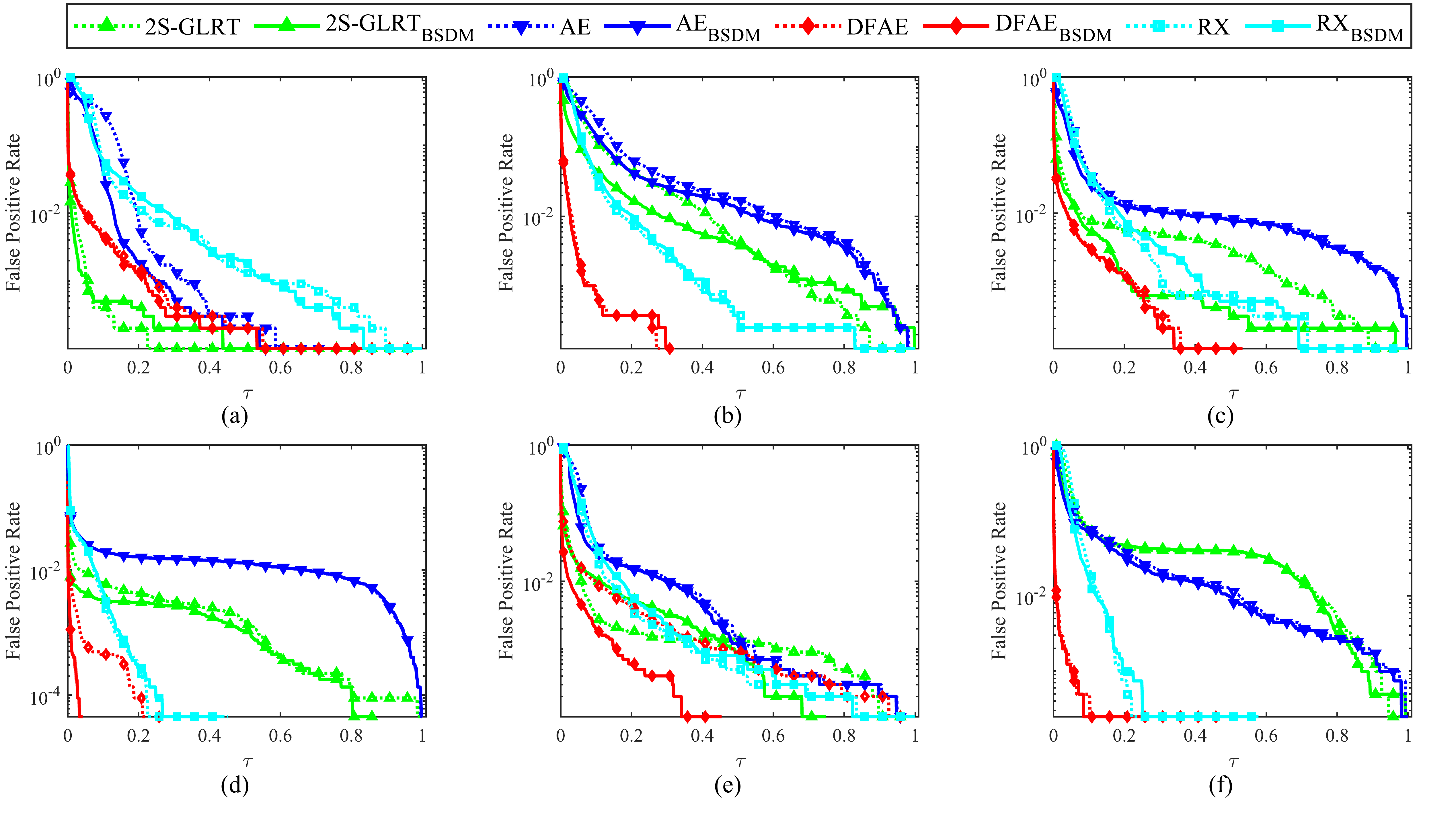}
	\caption{ROC curves of $(P_f,\tau)$ for different methods on (a) SanDiego, (b) Hydice, (c) Airport, (d) Beach, (e) Urban, (f) HAD-100.}
	\label{fig:f-roc}
\end{figure*}

\begin{table*}[ht]
	\centering
	\caption{Values of the Area Under the ROC $(P_d,P_f )$ and $(P_f,\tau)$ for RX on 6 Datasets. Results Using the Same Training Set and Test Set are Marked in Red. Baseline Denote the Result of Not Using BSDM. The Best Results are Bolded. Tr. and Te. Denote the Training Set and Test Set, Respectively}
	\begin{tabular}{c||cccccc||cccccc}
		\toprule[0.5mm]
		& \multicolumn{6}{c||}{\textbf{Values of the area under the ROC} $(P_d,P_f )$} & \multicolumn{6}{c}{\textbf{Values of the area under the ROC} $(P_f,\tau)$} \\
		\midrule
		\diagbox{\textbf{Tr.}}{\textbf{Te.}}  & \textbf{SanDiego} & \textbf{Hydice} & \textbf{Airport} & \textbf{Beach} & \textbf{Urban} & \textbf{HAD-100} & \textbf{SanDiego} & \textbf{Hydice} & \textbf{Airport} & \textbf{Beach} & \textbf{Urban} & \textbf{HAD-100} \\ \midrule
		\textbf{SanDiego} & \textcolor{red}{\textbf{0.9742}}  & \textbf{0.9985}  & 0.8865  & 0.9844  & \textbf{0.9658}  & 0.9803 & \textcolor{red}{0.0457}  & 0.0354  & 0.0361  & 0.0063  & 0.0335  & 0.0217 \\ 
		\textbf{Hydice} & 0.9716  & \textcolor{red}{0.9950}  & 0.8918  & \textbf{0.9869}  & 0.9620  & \textbf{0.9878} & 0.0438  & \textcolor{red}{\textbf{0.0274}}  & 0.0359  & \textbf{0.0056}  & \textbf{0.0312}  & \textbf{0.0188} \\ 
		\textbf{Airport} & 0.9620  & 0.9894  & \textcolor{red}{0.8814}  & 0.9856  & 0.9624  & 0.9803 & 0.0336  & 0.0333  & \textcolor{red}{\textbf{0.0336}}  & 0.0064  & 0.0342  & 0.0265 \\ 
		\textbf{Beach} & 0.9717  & 0.9880  & \textbf{0.8924}  & \textcolor{red}{0.9819}  & 0.9620  & 0.9805 & \textbf{0.0220}  & 0.0334  & 0.0370  & \textcolor{red}{0.0064}  & 0.0337  & 0.0233 \\ 
		\textbf{Urban} & 0.9658  & 0.9872  & 0.8833  & 0.9833  & \textcolor{red}{0.9655}  & 0.9813 & 0.0401  & 0.0283  & 0.0346  & 0.0065  & \textcolor{red}{0.0350}  & 0.0219 \\ 
		\textbf{HAD-100} & 0.9641  & 0.9951  & 0.8904  & 0.9824  & 0.9622  & \textcolor{red}{0.9801} & 0.0338  & 0.0325  & 0.0382  & 0.0065  & 0.0347  & \textcolor{red}{0.0342} \\ 
		\midrule
		\midrule
		\textbf{Baseline} & 0.9403 & 0.9763 & 0.8221 & 0.9808 & 0.9513 & 0.9797 & 0.0589 & 0.0380 & 0.0423 & 0.0065 & 0.0350 & 0.0455 \\ \bottomrule[0.5mm]
	\end{tabular}
	\label{tab:generalizability}
\end{table*}

\begin{table*}[!ht]
	\centering
	\caption{Values of the Area Under the ROC $(P_d,P_f )$ and $(P_f,\tau)$ for RX on 6 Datasets in the Case That Statistical Offset Module is Not Applicable. Baseline Denote the Result of Not Using BSDM. Tr. and Te. Denote the Training Set and Test Set, Respectively}
	\begin{tabular}{c||cccccc||cccccc}
		\toprule[0.5mm]
		& \multicolumn{6}{c||}{\textbf{Values of the area under the ROC} $(P_d,P_f )$} & \multicolumn{6}{c}{\textbf{Values of the area under the ROC} $(P_f,\tau)$} \\
		\midrule
		\diagbox{\textbf{Tr.}}{\textbf{Te.}}  & \textbf{SanDiego} & \textbf{Hydice} & \textbf{Airport} & \textbf{Beach} & \textbf{Urban} & \textbf{HAD-100} & \textbf{SanDiego} & \textbf{Hydice} & \textbf{Airport} & \textbf{Beach} & \textbf{Urban} & \textbf{HAD-100} \\ \midrule
		\textbf{SanDiego} & - & 0.9980  & 0.8833  & 0.9709  & 0.9612  & 0.9658 & - & 0.0427  & 0.0374  & 0.0063  & 0.0353  & 0.0241 \\ 
		\textbf{Hydice} & 0.9698  & - & 0.8955  & 0.9745  & 0.9601  & 0.9763 & 0.0463  & - & 0.0319  & 0.0060  & 0.0306  & 0.0195 \\ 
		\textbf{Airport} & 0.9618  & 0.9811  & - & 0.9803  & 0.9624  & 0.9712 & 0.0472  & 0.0403  & - & 0.0067  & 0.0371  & 0.0291 \\ 
		\textbf{Beach} & 0.9692  & 0.9859  & 0.8781  & - & 0.9560  & 0.9800 & 0.0296  & 0.0385  & 0.0332  & - & 0.0338  & 0.0222 \\ 
		\textbf{Urban} & 0.9635  & 0.9862  & 0.8807  & 0.9798  & - & 0.9663 & 0.0493  & 0.0401  & 0.0371  & 0.0066  & - & 0.0230 \\ 
		\textbf{HAD-100} & 0.9636  & 0.9932  & 0.8844  & 0.9785  & 0.9620  & - & 0.0374  & 0.0331  & 0.0665  & 0.0111  & 0.0449  & -\\ \midrule
		\midrule
		\textbf{Baseline} & 0.0589 & 0.0380 & 0.0423 & 0.0065 & 0.0350 & 0.0455 & 0.0589 & 0.0380 & 0.0423 & 0.0065 & 0.0350 & 0.0455 \\ \bottomrule[0.5mm]
	\end{tabular}
	\label{tab:generalizability-ablation}
\end{table*}

\subsection{Background Suppression Performances}
To validate the effectiveness of BSDM in background suppression, we analyze the anomaly-background separability by plotting the box plot in Fig. \ref{fig:box}. Columns such as RX refer to the anomaly-background separability results obtained by directly feeding the original HSIs into the corresponding detector. Columns such as RX$\rm{_{BSDM}}$ refer to the anomaly-background separability results obtained by feeding the original HSIs into the corresponding detector after background suppression with BSDM. Each detector corresponds to two boxes, where the red box indicates the distribution range of anomaly detection values and the green box indicates the distribution range of background class detection. The blue line in the middle of the box indicates the average value. The two ends of the black line outside the box indicate the maximum and minimum values. The position and compactness of the boxes indicate the trend of the background and anomaly pixels distribution. Therefore, after BSDM processing, the height of the green box decreases to indicate that the background is suppressed, and the greater the decrease, the stronger the background suppression. Meanwhile, the distance between the red box and the green box vertically indicates the degree of separation between anomalies and the background. As shown in the Fig. \ref{fig:box}, the height of the green box decreases and the distance between the red and green boxes and their averages increases in the detection results pre-processed by BSDM. This means that BSDM suppresses the background of the original HSIs and improves the separation of anomalies and background. 

Furthermore, by observing the visual comparisons in the Fig. \ref{fig:result-map}, we note that the background response is lower and the anomalies are more visible in the BSDM-processed result maps, which are reflected in both HSIs containing structured anomaly targets and HSIs with unstructured anomalous targets. Specifically, for HSIs containing structured anomalies, such as the airplane in Fig. \ref{fig:result-map} (a), RX can only detect the location of the airplanes from original image, while the shape of the airplanes is blurry and the upper right airplane with almost submerged in the background of airstrip. Conversely, after background suppression by BSDM, the shape of the airplanes in the RX detection result is more accurate, especially the one in the upper right, and the background of airstrip is darker, which indicates a lower response. For HSIs containing unstructured targets, such as the buildings in Fig. \ref{fig:result-map} (e), DFAE detects their locations but not obviously and there is a lot of background (upper right corner) that is falsely detected as anomalies. Conversely, after background suppression by BSDM, the anomalies in the DFAE detection results are more obvious, especially for the buildings located in the center of the image, and the background in upper right is mostly suppressed. In summary, BSDM can suppress the background of HSIs for a variety of scenarios, making anomalous targets more visible. 

To quantify the effectiveness of BSDM on HAD performance improvement, we tested the accuracy and false alarm rate of the HAD method before and after BSDM suppressed the background, as shown in the Tables \ref{tab:results-auc} and \ref{tab:results-fpr}, respectively. From Tables \ref{tab:results-auc} and \ref{tab:results-fpr}, we can observe that BSDM improves the performance of these HAD methods to different degrees. In particular, it improves the accuracy of 2S-GLRT on SanDiego by 26.39\%, while maintaining a low false alarm rate of 0.11\%. More notably, for methods with inherently better performance such as DFAE, BSDM can make further enhancements to them. For example, after BSDM inference, the accuracy of DFAE on Beach improved from 99.73\% to 99.94\%, and the false alarm rate decreased from 0.03\% to 0.01\%. As shown in the Fig. \ref{fig:result-map} and Fig. \ref{fig:box}, the proposed BSDM suppresses the background and improves the discrimination between the background and the anomalies. Meanwhile, the ROC curves of $(P_d,P_f)$ and $(P_f,\tau)$ are shown in Fig. \ref{fig:roc} and Fig. \ref{fig:f-roc}. Specifically, in the ROC curves of $(P_d,P_f)$, the BSDM shows higher $P_d$ as $P_f$ varying from 0 to 0.2. 

\subsection{Generalizability Study}
In this section, we take one of the six HSIs as the training set and the others as the test set to evaluate the generalizability of BSDM. To the best of our knowledge, the majority of deep learning-based HAD methods train and inference on a single dataset and do not explore generalizability, which means that generalizability verification in this paper is crucial. The few HAD methods with generalizability also have some problems, such as weaklyAD requires that the training and test sets are collected by the same sensor, and the method proposed in \cite{yoto} requires a large number of training sets. Importantly, BSDM has no specific requirements on the type of sensors (SanDiego and Hydice are collected by two different sensors) and the number of training sets (only one HSI is used as training set), which indicates better generalization and wider application scenarios. As shown in Table \ref{tab:generalizability}, although the training set is different from the test set, the proposed BSDM still improves the performance of the HAD methods. Also, the loss of performance is small compared to the case when the training set is the same as the test set (no more than 1.1\% drop in accuracy and no more than 0.8\% increase in false alarm rate). In particular, the accuracy and false alarm rates on HAD-100 are better when other HSIs are used as the training set than with itself as the training set. For example, when Hydice was used as the training set, the accuracy of HAD-100 improved by 0.77\% and the false alarm rate decreased by 1.54\%. This is because the band alignment strategy removes interference bands of HAD-100, which are generated during imaging. In summary, the proposed BSDM possesses well generalization, which means that after being trained, it only needs a few simple inferences to suppress the background of HSIs.

\subsection{Effectiveness of the Statistical Offset Module}
To explore the effect of the statistical offset module on the generalizability of BSDM, we tested the performance of BSDM without this module. The experimental results are shown in Table \ref{tab:generalizability-ablation}. In the case that statistical offset module is not applicable, BSDM maintains generalizability only on some data sets, such as SanDiego, Airport and Urban. While on the HAD-100 with a training set of SanDiego, the accuracy decrease by 1.43\% due to the excessive difference between the training and test sets. In the same case, the proposed statistical offset module improves the accuracy by 1.45\%. These results demonstrate that statistical offset module can indeed further improve the generalization of BSDM, which is consistent with the conclusion obtained in section \ref{sec:som}.

\section{Conclusion}
One key issue in HAD is how to develop a common method to suppress the background of HSIs for dealing with complex backgrounds, which has not been successfully resolved by existing HAD methods. We propose a background suppressed diffusion model for HAD called BSDM to enhance the discrimination of anomalies and background. To motivate the denoising network to learn the latent background distribution of the input HSIs, we designed a pseudo background noise which does not require any manual annotation. Considering the generalization of BSDM, we design a statistical offset module to make the denoising network adapt to datasets from different domains. Most importantly, we innovatively improve the inference process of DM by feeding the original HSIs into the denoising network, so that the background is removed as noise. Sufficient assessment and generalizability experiments demonstrate the effectiveness of BSDM in suppressing background and improving the performance of HAD methods.

\bibliography{ref}
\bibliographystyle{IEEEtran}

\end{document}